\documentclass[conference]{IEEEtran}
\IEEEoverridecommandlockouts
\usepackage{cite}
\usepackage{amsmath,amssymb,amsfonts}
\usepackage{algorithmic}
\usepackage{graphicx}
\usepackage{textcomp}
\usepackage{xcolor}
\def\BibTeX{{\rm B\kern-.05em{\sc i\kern-.025em b}\kern-.08em
    T\kern-.1667em\lower.7ex\hbox{E}\kern-.125emX}}
\begin{document}

\title{Creating Scalable AGI: the Open General Intelligence Framework\\

}

\author{\IEEEauthorblockN{1\textsuperscript{st} Daniel A Dollinger}
\IEEEauthorblockA{\textit{Technology Solutions Consulting} \\
\textit{NTT Data Ltd. North America}\\
Buffalo, New York, United States \\
0009-0004-4771-0613}
\and
\IEEEauthorblockN{2\textsuperscript{nd} Michael Singleton}
\IEEEauthorblockA{\textit{Technology Office} \\
\textit{Baker Street AI}\\
Valley View, Texas, United States \\
0009-0004-8429-6933}
}

\maketitle

\begin{abstract}
Recent advancements in Artificial Intelligence (AI), particularly with Large Language Models (LLMs), have led to significant progress in narrow tasks such as image classification, language translation, coding, and writing. However, these models face limitations in reliability and scalability due to their siloed architectures, which are designed to handle only one data modality (data type) at a time. This single-modal approach hinders their ability to integrate the complex set of data points required for real-world challenges and problem-solving tasks like medical diagnosis, quality assurance, equipment troubleshooting, and financial decision-making. Addressing these real-world challenges requires a more capable Artificial General Intelligence (AGI) system.

Our primary contribution is the development of the Open General Intelligence (OGI) framework, a novel systems architecture that serves as a macro design reference for AGI. The OGI framework adopts a modular approach to the design of intelligent systems, based on the premise that cognition must occur across multiple specialized modules that can seamlessly operate as a single system. OGI integrates these modules using a dynamic processing system and a fabric interconnect, enabling real-time adaptability, multi-modal integration, and scalable processing.

The OGI framework consists of three key components: (1) Overall Macro Design Guidance that directs operational design and processing, (2) a Dynamic Processing System that controls routing, primary goals, instructions, and weighting, and (3) Framework Areas, a set of specialized modules that operate cohesively to form a unified cognitive system. By incorporating known principles from human cognition into AI systems, the OGI framework aims to overcome the challenges observed in today's intelligent systems, paving the way for more holistic and context-aware problem-solving capabilities.

\end{abstract}


\section{Introduction}\label{section1}
The most recent Artificial Intelligence (AI) breakthroughs with Large Language Models (LLMs) has led to many advancements in the way AI is used and applied in everyday processes.  With these advancements have come many benefits for a broad number of narrow tasks such as image classification, language translation, coding, and writing.  Unfortunately, much of this advancement is still plagued by limitations in reliability which manifest themselves in common occurrences such as hallucinations. 

Current AI models face significant limitations due to their architecture. They are typically designed to handle only one type of data (e.g. text, images, audio, etc) at a time and operate within isolated frameworks.  As a result, they struggle to integrate different types of data, which is required for building a broad enough understanding to solve problems effectively. This issue is not solvable by simply increasing computer power.  A common example is medical diagnostics, combining patient history (text), lab results (numeric data), and medical images (visual).  A model limited to only one modality (data type) will miss critical information.

A more holistic architecture that can handle a broader set of data modalities (natively) is required.  To address these current challenges, look to the human brain for inspiration.  This paper’s intention is to incorporate known principles from human cognition into artificial intelligence systems.  

The proposed architecture, open general intelligence framework (OGI), is intended to be a macro design reference for general intelligence as defined by common known human cognition capabilities.  The OGI architecture differs from existing methods through real-time adaptability, multi-modal integration, and scalable processing.  Unlike traditional static AI models, OGI’s dynamic system adjusts tasks and resource allocation while specialized processing modules collaborate seamlessly to process diverse data types.  With additional capabilities such as cognitive process switching and an interconnected processing fabric, OGI represents a reference architecture for artificial general intelligence that mimics human-like cognitive flexibility, addressing complex and able to tackle real-world challenges with greater contextual awareness and efficiency. 

For clarity, OGI is not intending to replicate the human brain; rather, OGI is identifying key traits and operational processes that are believed to be present in general intelligence. The architecture consists of three distinct tenants:

\begin{enumerate}
    \item \textbf{Overall Macro Design Guidance} that guide operational design and processing
    \item \textbf{Dynamic Processing System} that controls routing, primary goals, instructions, and weighting
    \item \textbf{Modular Architecture Areas} distributed across specialized functional modules that operate as one system
\end{enumerate}

The OGI framework has been outlined below into macro guidance, control, and areas below:

\textbf{Framework Macro Design Guidance}
\begin{enumerate}
    \item Multiple Data Type Support
    \item Multiple Specialized Processing Modules
    \item Interconnected Processing Fabric
    \item Cognitive Process Switching
\end{enumerate}

\textbf{Framework Control}
\begin{enumerate}
    \item Dynamic Processing System
\end{enumerate}

\textbf{Framework Areas}
\begin{enumerate}
    \item Executive Control
    \item Autonomous Processing 
    \item Input/Output Integration
    \item Short Term Memory
    \item Long Term Memory
    \item Fabric Interconnect
\end{enumerate}

\begin{figure}[h!]
    \centering
    \includegraphics[width=1\linewidth]{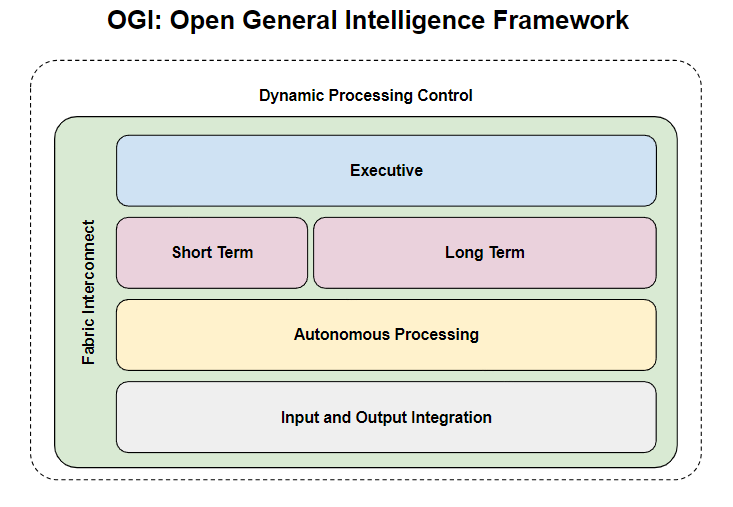}
    \caption{The OGI Framework Architecture}
    \label{fig:OGIF}
\end{figure}

The intended structure of this framework is not linear, with several areas such as short and long term memory having overlap. Furthermore, each area will have mesh connections provided through the fabric interconnect.  For a visual representation, please refer to the architecture diagram (figure \ref{fig:OGIF}) in section \ref{section3}.

\section{Background: Human Cognition as Inspiration}
More recent AI breakthroughs in computational scaling have brought us to the current moment where AI is proving enormous amounts of broad-market utility and can solve real world problems at scale.  Though promising, AI such as LLMs have proven themselves to be narrow aligned in real-world applications.  This narrow focus is misaligned with what we know of the human brain’s broader cognitive capabilities, which we seek to address in the proposed OGI architecture. 

The human brain’s native neural processing is able to adjust and assimilate new information on the fly from both internal and external sources, building both macro and micro context to aid in more rigorous decisioning \cite{b1}.  When it builds this context, it is assembled together into coherent thought patterns that have multi-dimensional attributes such as space, time, human senses, past memories, social context, and emotions.  As these attributes are unpacked, it becomes apparent that single modality data processing in AI models is primitive compared to the human brain.  
The gap of today’s single-modality AIs such as LLMS becomes readily apparent through some examples that highlight multiple data types being required.  The following examples highlight the complexity that single-modalities are unable to solve:
\begin{itemize}
    \item Medical diagnosis requires patient history (text), lab results (numeric), examination (physical touch and visual), and images (visual).  
    \item Sarcasm and irony requires tone (audio), facial expression (visual), broader situational cues (text, audio, visual).
    \item Quality assurance requires aesthetics (visual and tactile), functionality (multiple), and customer appeal (emotional).
    \item Safety requires inspection (visual), cost (numeric), structural or environment integrity (physics), and legal compliance (text and regulatory context).
    \item Financial decision making requires economic forecasts (statistics), compliance changes (text and regulatory context), public sentiment (macro emotional), company strategy (spacial time and planning).
    \item Technology and equipment troubleshooting requires listening (auditory), feeling vibrations (tactile), seeing (visual), reviewing maintenance history (text), and evaluating surrounding facts such as environment and overall quality of outputs (multiple).
\end{itemize}

\begin{figure}
    \centering
    \includegraphics[width=1\linewidth]{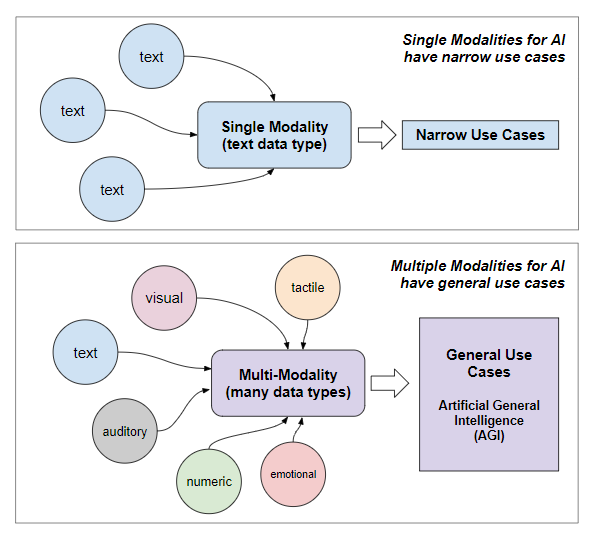}
    \caption{Multiple data types allows for artificial general intelligence}
    \label{fig:modality-comparison}
\end{figure}

Comparing an AI such as an LLM to the human brain may not be a fair comparison.  It is well established that the human brain is made up of distinct and interconnected modules that specialize in different processing capabilities.  Furthermore, these modules work together in a series of both autonomous and learned thought process frameworks to solve problems.  In example, a mathematics or language framework can be instilled and be habitualized into a thought pattern that becomes an automated cognitive process across brain modules.  Though some brain modules do indeed specialize in types of processing, responsibilities are shared to create coherent thought patterns.  This macro view of thinking reflects some key requirements for any AI system attempting to move beyond narrow tasks to broader processing:

\textbf{Intelligent System Macro Design Guidance}
\begin{enumerate}
    \item Multiple Data Type Support
    \item Multiple Specialized Processing Modules
    \item Interconnected Processing Fabric
    \item Cognitive Switching between Automated and Logical Processing
    \item Controllable Context Switching
\end{enumerate}
These macro requirements have been broken out in detail below.

\subsection{Multiple Data Type Support}

The human brain relies on multiple input data types to build context, allowing both autonomous and manual decisions to be processed as outputs.  These inputs take the form of both internal and external data points that can range from simplistic feelings to complex aggregations of multiple inputs \cite{b2}.

Internal sources can take the form of memories, emotional responses, hormonal, cognitive directives, bias, and more complex combinations of inputs.  This list is not exhaustive, and there is often some ambiguity between what defines and input and output as there can be multiple parallel inputs and outputs that are intertwined (e.g. hormonal output is an input to both emotions and temperature increase, which become inputs to a cognitive decision).

External sources enter via a range of methods.  These inputs are spread across many different sensory mediums, ranging from the standard human five senses to more abstract sources such as social context.  In the same manner as internal inputs, external inputs may also have complex combinations, such as sight, touch, smell, and social context. (e.g. sight, feeling, and auditory are combined to create “hot fire nearby”).

\subsection{Multiple Specialized Processing Modules}

The human brain’s modular architecture consists of multiple specialized processing modules that work together in a coordinated manner to break down problems for complex decisions.  In example, the inferior frontal gyrus is believed to house language while the occipital lobe near the back houses visual  \cite{b3,b4}.  

Having multiple modules allows for processing of different data types efficiently, similar to how hardware offloading in a computer delegates tasks to an ASIC (application specific integrated circuit).  

This expands beyond efficiency and is an intrinsic property in the human brain.  In example, visual data is processed by the primary visual cortex in the back of the brain.  Once processed, it is able to be combined with data from other sensory modules and internal sources (e.g. memory) to build a broader context not possible with visual processing alone \cite{b5}.  

\subsection{Interconnected Processing Fabric}

In order to process data across modules, the human brain has a fabric of connections described as neural pathways and synaptic connections \cite{b6}.  This fabric is able to pass information across modules, and evidence even shows that it plays an active role in transforming data in transit \cite{b7}.  Arguably, the capabilities of the brain’s interconnected fabric may be one of the most intriguing problems to reproduce in any engineering framework.

When we consider how data is processed across brain modules, there are several mechanisms that stand out.  First is the ability for the fabric to reorganize how data is processed depending on what is needed \cite{b8}.  Second, the brain uses feedback and forward feedback to refine and combine multiple data types across modules \cite{b4}.  Third, the brain performs multiple tasks in parallel \cite{b9}.

\subsection{Cognitive Process Switching}

A key tenant of human cognitive processing is the transitory processes that allow for switching between different cognitive processing strategies.  By default, the brain follows learned strategies for working through challenges.  These automated patterns allow the brain to seamlessly utilize different parts of the brain depending on the type of challenge (e.g. language, math, motor, etc.) \cite{b10}.

When the brain is unable to seamlessly solve a challenge, it transitions to a logical (manual) state to approach the challenge through logic \cite{b10}.  In example, performing a mathematical calculation in one’s head. Switching between autonomous and logical states occurs daily, and in many cases these states continue to operate in parallel. 

When an automated routine or habit initiates,  autonomous processing takes over to operate until interrupted.  The logical process is then freed up to focus on more priority tasks such as thinking through another future task \cite{b10}.  In example, a person may embark on a regular walk on a trail.  As they walk, their logical process reflects on work or relationship challenges. Meanwhile, their muscles and senses autonomously take them on the habitual journey.  During this time, the brain matches the incoming context created by incoming environment sensory data with the internal context stored from memory.  As long as both of these contexts generally match, an interrupt does not happen \cite{b4,b10}.  As soon as something out of context occurs, such as a fallen tree in the path, the logical context receives an interrupt to take over.

\section{Proposed Intelligent System Architecture}\label{section3}

The goal of the proposed architecture is to emulate human cognitive processes to allow for generalized AI systems that are scalable and reliable.  The proposed reference architecture emulates key human cognition capabilities using a series of functional processing areas.  This is required based on the variety of data required to decode all the data points typically seen in real world decision making.

This standardized architecture will allow for the creation of AGI that can be used across a variety of real-world applications such as medical diagnosis, quality assurance, environmental inspections, financial decision making, legal frameworks, equipment diagnostics, and even engineering design.

The OGI framework has been outlined below:

\textbf{Framework Macro Design Guidance}
\begin{enumerate}
    \item Multiple Data Type Support
    \item Multiple Specialized Processing Modules
    \item Interconnected Processing Fabric
    \item Cognitive Process Switching
\end{enumerate}

\textbf{Framework Control}
\begin{enumerate}
    \item Dynamic Processing System
\end{enumerate}

\textbf{Framework Areas}
\begin{enumerate}
    \item Executive Control
    \item Autonomous Processing 
    \item Input/Output Integration
    \item Short Term Memory
    \item Long Term Memory
    \item Fabric Interconnect
\end{enumerate}

The intended structure of this framework is not linear, with several areas such as short and long term memory having overlap. Furthermore, each area will have mesh connections provided through the fabric interconnect.

Each of the area’s functional capabilities have been described, with special consideration for the all requirements outlined in each of the following subsections.  This framework is not intended to be linear. The OGI framework is outlined below:

\subsection{Intelligent System Macro Design Guidance}

The intelligent system’s macro capabilities describe overall design guidance when building an intelligent cognitive system.  
\begin{enumerate}
    \item \textbf{Multiple Data Type Support} - The intelligent system should support multiple data types to allow for multi-dimensional context generation and cognitive processing.
    \item \textbf{Multiple Specialized Processing Modules} - The intelligent system should utilize multiple processing modules that accelerate specialized cognitive processing by area.
    \item \textbf{Interconnected Processing Fabric} - The intelligent system should interconnect modules and provide real-time cognitive processing across modules.
    \item \textbf{Cognitive Process Switching}  - The intelligent system’s cognition processes should allow for automated routing to the right processing area, switching between automatic and logical processing to increase efficiency, reduce cognitive load, and escalate to more intelligent processing as required.
\end{enumerate}

\subsection{Framework Control: A Dynamic Processing System}
 
Scaling, reliability, and safety are arguably the largest challenges in the AI field.  Neither of these are solvable solely through scaling out computation power.  It is necessary to move beyond this siloed mentality and design with efficiency and reliability as core tenants.  Scaling and reliability is achieved by aligning problems and data types with the best processing architecture.  Similar to the brain, processing should be distributed across specialized modules.

Moving to distribute processing across modules will require a dynamic processing system that coordinates processing across modules in a coherent fashion.  It will need to operate similar to an ASIC, routing processes to their locations depending on the type of challenge or data type. 

The implications are that embedded inside the dynamic processing system should be programmable instruction layers that control routing, primary goals, instructions, and weights.  This programming area should be able to set the following:

\begin{enumerate}
    \item Routing adjustments that control processing of the intelligent system
    \item Primary goals and instructions that set the overall objective and focus of the intelligent system
    \item Weights that adjust context and how the system approaches tasks
\end{enumerate}

These programmable instruction layers may have different mechanisms depending on cognition area, but should operate in concert with each other to steer towards the primary goal and tune towards related tasks.  In example, a primary goal may be to perform research on cats.  The programmable layers should shift routing, context and weights to be more logical as opposed to creative.  

For control and safety, there should be an external programming administration area to adjust the primary goal and instructions of the overall cognition of the intelligent system.  This will maintain control and safety.  

\begin{figure}[h]
    \centering
    \includegraphics[width=1\linewidth]{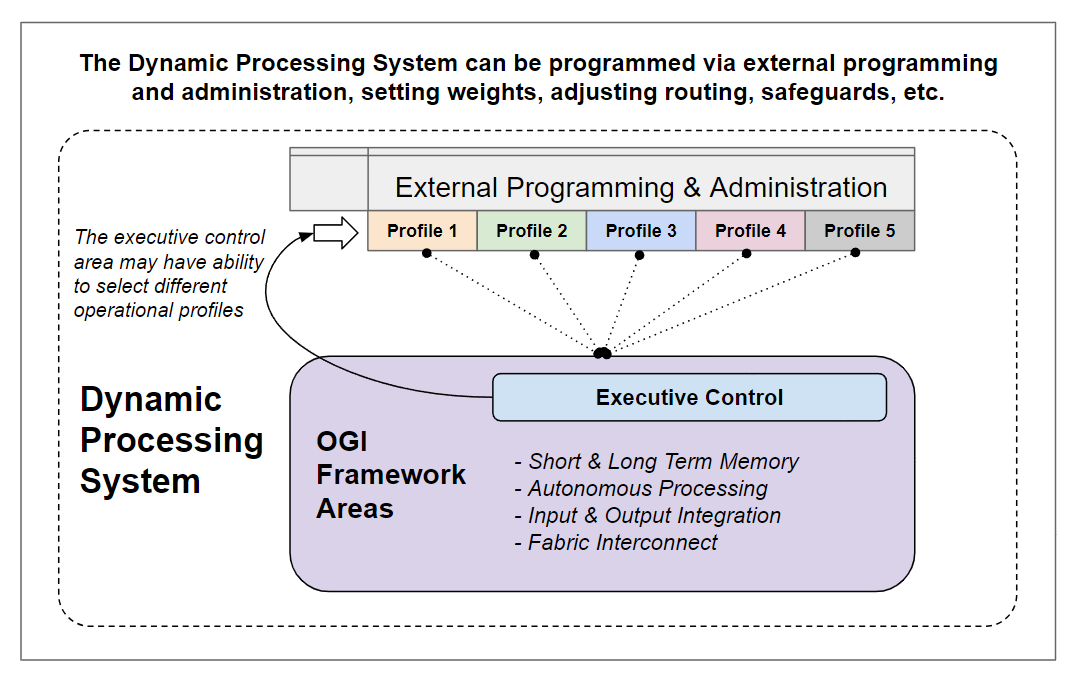}
    \caption{The intelligent system can be programmed for dynamic operations}
    \label{fig:dynamic_processing}
\end{figure}

Internal to the intelligent system, there should be a limited set of controls that allows the executive control area to adjust settings such as the weights across portions of the intelligent system based on the current task.  This could potentially be achieved by allowing the executive control area access to select different operational profiles depending on the task at hand (e.g. logical, creative, motor).  However, the goal should be for minimal touch  This balanced approach will provide both programmable guardrails and a level of autonomy to operate. 

\subsection{Framework Area: Executive Control}

The executive control area functions at the highest level of the AI brain,  providing an internal monologue of oversight and  generalized logical reasoning. As the central reasoning model, it is important to note that this is a proactive AI that is constantly monitoring area statuses through short term memory’s context.  As it monitors, it reflects on next decisions, reviews past experience, and determines the best way to achieve objectives.

The AI receives its instructions through the dynamic weighting framework, which as previously discussed, has two layers, an external and internal.  The external layer is programmed external to the cognition system.  This sets the primary goal and instructions on how to operate.  The executive control area will have access to update the internal layer of the dynamic weighting framework in order to optimize its cognitive processes as it works through challenges.

Processing for the executive control area takes place in short term memory.  As a working space, this area serves as a staging ground for current thought monologue as it interacts with the current state context of the broader AI brain system.  This context can be used to reason and solve complex problems in real time, providing the ability for the context to be updated.  For example, if the AI system is perceiving a scenario where it must make a decision, the associated context in short term memory should make a connection to long term memory where it can use that as reference to guide a decision.  Therefore, the internal monologue evaluates and makes an informed decision. 

The cognition internal to the executive control area should be weighted towards being a generalized model for stability and flexibility.  If more specialized training is required for this model, balancing between generalized and specialization may be achieved with the addition of methods such as retrieval augmented generation (RAG) through input and output integration (see Input and Output Integration below).  Furthermore, the autonomous control layer may be a specialized model based on the types of inputs and outputs it should be controlling.

\subsection{Framework Area: Autonomous Processing Area}

The autonomous processing area functions as the core AI execution, coordination, and context reporting  model.  Its autonomous nature allows it to function quickly and reactively without much assistance from the executive control area.  Much of how it functions is similar to a series of stored procedures embedded inside a multi-modal model.

As the autonomous area  executes and coordinates outputs, it processes data across modalities in real time.  This allows it to match incoming context to previously learned stored procedures.  In example, pedaling a bicycle can be done effortlessly once the motor outputs are placed into the context of riding a bicycle.  While under execution, these stored procedures may be looped (inputs and outputs) until the current context changes or ends.  In the same example, as the bicycle rider reaches the end of the path, the situational context changes and the stored procedure of pedaling is interrupted.

\begin{figure}[h]
    \centering
    \includegraphics[width=1\linewidth]{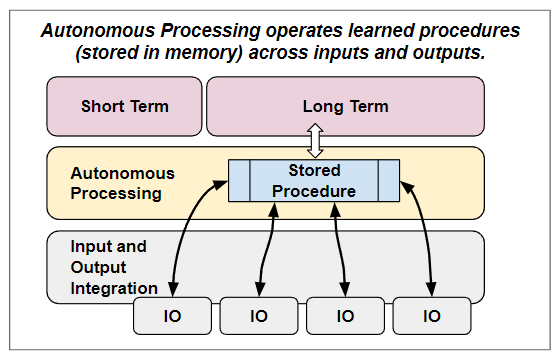}
    \caption{Autonomous processing operates with minimal executive assistance}
    \label{fig:stored_procedure}
\end{figure}

Whereas the autonomous processing area can operate standalone, it is required to constantly report status in short term memory to the execution layer in the form of current state context.  This context provides what can be described as  perception, made up of sensory inputs and current status of processing.  The executive control area does not have direct visibility into the autonomous processing area without short term memory.  This is a one-way relationship as the autonomous area has no visibility into the executive area.

\subsection{Framework Area: Input and Output Integration}

The input and output (IO) integration area describes the way all IOs  interface across different areas.  Each IO type will have its own associated model that most effectively processes its modality type.  Examples of inputs include various sensors, databases, and files from storage.  Examples of outputs include spoken language, motor control, and image generation. 

In order to integrate effectively across areas, each will require a consistent means of communications and controls.  This is how an application programming interface (API) works, where the benefit is that separate and distinctly different entities can interface with one another.

Building a standardized IO integration layer allows for modular expandability of the overall proposed architecture, allowing additional IOs to be layered in to minimize redesigns of the executive and autonomous areas. 

Both the executive and autonomous areas will have control over inputs and outputs via the API-like capabilities.  The autonomous area will have a significant advantage over the executive area in terms of reaction speed and coordination across IOs due to its stored procedure capabilities.  However, the executive area will have distinct advantages when attempting to control unique and new actions. 

Note that in the case that some IOs require their own specialized model (e.g. image generation or LLMs), these models will interoperate with the broader intelligent system through the standard IO integration area.  

\subsection{Framework Area: Short Term Memory}

As a working space for information and context, short term memory is a temporary space to process context, report status, and make decisions.  Similar to a computer’s non-volatile random access memory (NVRAM), short term memory is highly performant and the data contained is loaded in as required.  It stays persistent across system resets, though its finite space will require it to trim data or store it in long term memory.

Both the executive control area and autonomous processing area utilize short term memory as a working space for creating current context.  The autonomous processing layer uses short term memory to maintain operational continuity across stored procedures and for reporting to the executive area.

The executive area uses short term memory as a working space for operations as well as understanding context generated by the autonomous area.  As an executive working space, complex decisions can be made, imaginative processes can generate digital representations, context can be updated, and updates to long term memory can eventually take place.

\subsection{Framework Area: Long Term Memory}

As a more permanent storage place for information, long term memory services as a place to reference how to operate in the future and make decisions.  Though in concept similar to a computer’s permanent solid state storage, it differs in the way that memory itself may be a spectrum between short and long term.  This implies there may be no clear transition phase between short term and long term, and the key difference is long term memory is less likely to be forgotten based on how strong its connections are.

\begin{figure}[h]
    \centering
    \includegraphics[width=1\linewidth]{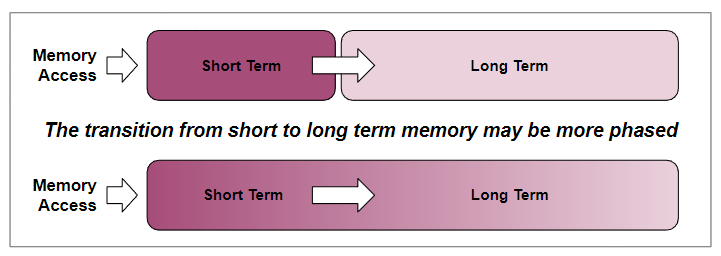}
    \caption{The transition from short to long term memory may be phased}
    \label{fig:memory-transition}
\end{figure}

Long term memory is accessed through short term memory, which relies on context to create connections to long term information.  Over time, as long term memory is referenced, its connections and persistence to particular types of data may increase.  This model assumes that memory itself must be pliable and learning is required to happen throughout the lifespan of the AI brain.

\subsection{ Framework Area: Fabric Interconnect}

A flexible and highly performant connection network is required to facilitate fast and seamless communication between brain modules.  This multi-pathing fabric interconnect will be required to support simultaneous transfer of information across multiple modules.  

As a many-to-many network, near zero latency will also require hardware level processing speeds that can facilitate intra-neural communications between modules.  In the case of a lower-performant fabric interconnect, a queuing system will be required.  This will come at the expense of reactionary speeds for IO and potentially result in incorrectly generated context as order could matter in some scenarios (e.g. movement coordination).

\section{Known Challenges and Future Considerations}

The implementation of OGI presents several challenges and limitations that must be addressed to realize its full potential. These have been broken out by challenge below:

\subsection{Controlling Weights in Real Time}
 Current AI models rely on adjustments that are often manually optimized for narrow use cases.  These are often meticulously trialed and tuned to ensure optimum results and must be revisited as use cases update.  In order to move towards general AI, establishing an effective control mechanism to balance and prioritize the different processing modules is crucial. Dynamically processing adjustments in real time with limited to no latency will require ASIC-like performance.
 
The dynamic weighting framework can be mathematically represented as:
 
\begin{equation}
\label{eqn_example1}
\Phi: (C, E_t) \rightarrow \Delta^n 
\end{equation}
where
\begin{equation}
\label{eqn_example2}
w_t = \Phi(C, E_t) = \text{softmax}(g(C, E_t))
\end{equation}

This raises questions about how weights are re-calibrated in real-time and what guiding principles or objective functions optimize this weighting. Developing a robust and flexible weighting system that can adaptively coordinate various specialized modules remains a significant technical hurdle, as highlighted by recent studies on dynamic cognitive architectures \cite{b11}.  

\subsection{Coordinating Cognition Across Modules}

 Integrating specialized processing modules (e.g., visual, linguistic, memory) into a cohesive cognitive system poses substantial challenges. Ensuring smooth information flow and coordinated decision-making requires sophisticated mechanisms for communication and conflict resolution among modules.  Achieving human-like cognitive fluidity across diverse processing capabilities is a major frontier in AI research \cite{b12}.
 
Practically, this will require accelerated transport layer networking for distributed systems, and for closed systems, high performance communication lanes. This transport layer will require low latency, bi-directional protocols that can be accelerated with ASICs and route information between modules in real time.  Moving information control to higher layers could potentially achieved with messaging protocols for slower and more calculated cognitive tasks.

\subsection{Multi-Modal Processing }
The ability to seamlessly integrate and reason across multiple data modalities (vision, language, sensory input) is a hallmark of human cognition that current AI systems struggle to replicate. Supporting heterogeneous input/output types necessitates developing models capable of understanding and reasoning about multimodal information effectively. Techniques such as attention mechanisms may refine unimodal representations but scaling this understanding to achieve human-level flexibility remains an open problem \cite{b13}.

An example of the scale of the challenge can be illustrated with smell.  The scent of a favorite food can instantly connect human cognition to generate images, tastes, sounds, and long term memories in real time, resulting in autonomous outputs such as salivation and hunger pangs.  This example implies that there are intrinsic connections between different data modalities in memory.  These multimodal associations provide broader context through more data points \cite{b5}.  The autonomous processing area and IO integration area will both require bidirectional and feedback mechanisms to enable congruent associations.

\subsection{Combining Learning and Training}
Current AI systems often rely on rigid, batch-oriented training methods, which differ from the fluid, integrated memory mechanisms observed in human cognition. While OGI implements multiple interacting memory processes, the specifics of how these mechanisms cooperate and consolidate information over different timescales need further elaboration. The challenge lies not in separating memory types, but in understanding how different memory mechanisms work together dynamically.  This is similar to how biological memory systems operate as a continuous, interactive process rather than discrete stores. Achieving adaptive, continually learning AI systems that exhibit this kind of integrated memory processing remains an elusive goal that OGI aims to address.

\subsection{Future Considerations}

Future research directions for OGI must address both theoretical foundations and practical implementation challenges. From a theoretical perspective, developing objective functions for $g(C, E_t)$ that optimally balance module contributions is crucial, alongside investigating meta-learning approaches to automatically adapt the weighting function $\Phi$ based on task performance. This mathematical foundation must be complemented by exploring theoretical bounds on convergence properties and establishing formal guarantees for system stability under module reconfiguration. Additionally, extending the $\Delta^n$ simplex representation to handle hierarchical module relationships will be essential for scaling the architecture to more complex cognitive tasks.

The practical advancement of OGI requires significant developments in multi-modal integration and empirical validation. Research priorities include developing efficient attention mechanisms that can scale across increasing numbers of specialized modules, alongside information-theoretic metrics for measuring cross-module coordination efficiency  \cite{b14}. The empirical validation framework must include benchmark tasks specifically targeting the dynamic processing system's adaptation speed, standardized metrics for measuring cognitive fluidity across modules, and established baselines for module coordination overhead and resource utilization. These practical advances should be guided by theoretical insights from optimal control theory and analysis of the combined learning-weighting dynamics.

Looking beyond current capabilities, OGI must evolve to handle increasingly complex real-world scenarios. This evolution requires investigating reinforcement learning techniques that allow the system to learn from environmental interactions dynamically, while incorporating probabilistic models to enhance uncertainty management in decision-making processes. Future research should systematically evaluate OGI against existing models across diverse tasks and datasets, with particular attention to the architecture's ability to maintain stability while adapting to novel situations. The ultimate goal remains developing a cognitive framework that combines theoretical rigor with practical adaptability, capable of approaching human-like flexibility in real-world applications.

\section{Validation in Practice}

Validating the effectiveness and reliability of the proposed OGI architecture requires a multi-faceted approach, addressing each component area as well as the overall system performance. 

\subsection{Benchmarks and Metrics}

First, its ability to handle diverse real-world tasks and data modalities must be assessed. While no single benchmark perfectly captures OGI’s capabilities, we propose adapting existing datasets like ImageNet \cite{b15} and COCO \cite{b16} by augmenting them with synthetic audio or textual labels, forcing OGI to integrate modalities for optimal performance. Additionally, OGI will be evaluated on modified multi-modal benchmarks, such as those used for Visual Question Answering \cite{b17}. Key metrics will include accuracy, efficiency, and the ability to outperform unimodal models or naive fusion methods. To assess generalization, OGI will be trained on one dataset and tested on a different but related one, with novel stimuli introduced during testing. The drop in accuracy compared to standard models will demonstrate GOI’s robustness to out-of-distribution data \cite{b18}.

\subsection{Internal Monitoring}

Further validation will focus on analyzing OGI’s internal decision-making processes. Extensive instrumentation and monitoring will trace information flow, observe module activations, and understand how the dynamic weighting system adjusts over time. This will involve designing specific tasks, such as those requiring rapid task switching \cite{b19} or dynamic resource allocation \cite{b20}, to assess the Executive Control module's effectiveness. Key metrics will include time taken to switch tasks, accuracy under varying cognitive load, and efficient resource utilization. Finally, scalability and efficiency will be evaluated by varying the complexity of tasks and testing OGI on different hardware platforms.  Processing time, memory usage, and energy consumption will be measured as the system scales. By rigorously evaluating OGI across these dimensions, we can build confidence in its potential to address limitations of current AI systems and advance towards more general and adaptable intelligence.

\subsection{Practical Validation}

Beyond benchmark performance, the ultimate validation of OGI lies in its ability to perform complex tasks in real-world environments with minimal human intervention. This requires evaluating OGI in situated scenarios, such as those encountered in robotics \cite{b21}, human-computer interaction\cite{b22}, or autonomous systems. These evaluations will assess OGI’s capacity to integrate multi-modal information, adapt to dynamic and unpredictable situations, and make effective decisions with limited human guidance. Key metrics will include task completion rate, efficiency of resource utilization, and the ability to handle unexpected events or changes in the environment. Furthermore, scalability will be assessed by deploying OGI on increasingly complex real-world tasks, measuring its performance and resource consumption as the scale and complexity increase.

As OGI matures towards real-world acceptance, the true test of validation will be industry type certifications.  These certifications may be based on existing human tests such as medical exams; however, new certification methods will likely need to be developed.  LLMs already can pass many human certification tests, but in reality, they fail in real world scenarios as they lack the ability to holistically process outside a closed system.  New certification tests will need to approach problems more holistically, applying multiple methods of rigor that demonstrate the ability for the intelligent system to process all methods of data as well as reliability work through challenges in a way that provides the intended outcomes.

\section*{Acknowledgment}

The authors would like to express sincere respect and gratitude to those who have poured their lives work into research in the fields of human cognition and artificial intelligence.  This framework stands on the "shoulders of giants," as it would not have been possible without the foundational contributions of these pioneers.  Special thanks to colleagues and peers who have provided constructive feedback and insightful discussions during the refinement phases of this work.  The author would also like to thank the IEEE community for providing access to invaluable resources that contributed to making this possible.


\begin{thebibliography}{00}

\bibitem{b1} 
M. Chawla and K. P. Miyapuram, "Context-Sensitive Computational Mechanisms of Decision Making," \textit{Journal of Experimental Neuroscience}, vol. 12, Nov. 2018, Art. no. 1179069518809057. doi: 10.1177/1179069518809057

\bibitem{b2}
Okazawa, Gouki, and Roozbeh Kiani. “Neural Mechanisms That Make Perceptual Decisions Flexible.” Annual review of physiology vol. 85 (2023): 191-215. doi:10.1146/annurev-physiol-031722-024731

G. Okazawa and R. Kiani, "Neural Mechanisms That Make Perceptual Decisions Flexible," \textit{Annual Review of Physiology}, vol. 85, pp. 191-215, 2023. doi: 10.1146/annurev-physiol-031722-024731

\bibitem{b3} 
Z. J. Chen, et al., "Revealing Modular Architecture of Human Brain Structural Networks by Using Cortical Thickness from MRI," \textit{Cerebral Cortex}, vol. 18, no. 10, pp. 2374-2381, 2008. doi: 10.1093/cercor/bhn003

\bibitem{b4} 
S. Sternberg, "Modular processes in mind and brain," \textit{Cognitive Neuropsychology}, vol. 28, no. 3-4, pp. 156-208, 2011. doi: 10.1080/02643294.2011.557231

\bibitem{b5} 
C. M. A. Pennartz, et al., "How 'visual' is the visual cortex? The interactions between the visual cortex and other sensory, motivational and motor systems as enabling factors for visual perception," \textit{Philosophical Transactions of the Royal Society of London. Series B, Biological Sciences}, vol. 378, no. 1886, p. 20220336, 2023. doi: 10.1098/rstb.2022.0336

\bibitem{b6} 
M. B. Kennedy, "Synaptic Signaling in Learning and Memory," \textit{Cold Spring Harbor Perspectives in Biology}, vol. 8, no. 2, p. a016824, Dec. 2013. doi: 10.1101/cshperspect.a016824

\bibitem{b7} 
L. C. Sincich, et al., "Preserving information in neural transmission," \textit{The Journal of Neuroscience: The Official Journal of the Society for Neuroscience}, vol. 29, no. 19, pp. 6207-6216, 2009. doi: 10.1523/JNEUROSCI.3701-08.2009

\bibitem{b8}
D. Vatansever, et al., "Default Mode Dynamics for Global Functional Integration," \textit{The Journal of Neuroscience: The Official Journal of the Society for Neuroscience}, vol. 35, no. 46, pp. 15254-15262, 2015. doi: 10.1523/JNEUROSCI.2135-15.2015

\bibitem{b9}
E. J. Müller, et al., "Parallel processing relies on a distributed, low-dimensional cortico-cerebellar architecture," \textit{Network Neuroscience}, vol. 7, no. 2, pp. 844-863, Jun. 2023. doi: 10.1162/netn{\_}a{\_}00308

\bibitem{b10} 
A. W. Sali, et al., "Learning Cognitive Flexibility: Neural Substrates of Adapting Switch-Readiness to Time-Varying Demands," *Journal of Cognitive Neuroscience*, vol. 36, no. 2, pp. 377-393, 2024.


\bibitem{b11}
 J. Laird and P. Derbinsky, "A Rational Meta-Cognitive Architecture," Artificial Intelligence, vol. 216, pp. 53-79, 2014.
 
\bibitem{b12} 
K. Fukushima and Y. Miyake, "Neocognitron: A Self-Organizing Neural Network Model for a Mechanism of Pattern Recognition Unaffected by Shift in Position," Biological Cybernetics, vol. 36, no. 4, pp. 193-202, 1980.

\bibitem{b13}
Y. LeCun, Y. Bengio, and G. Hinton, "Deep Learning," Nature, vol. 521, no. 7553, pp. 436-444, 2015.

\bibitem{b14} 
A. Vaswani et al., "Attention Is All You Need," Advances in Neural Information Processing Systems, vol. 30, pp. 5998-6008, 2017.

\bibitem{b15}
J. Deng, W. Dong, R. Socher, L.-J. Li, K. Li, and L. Fei-Fei, "ImageNet: A large-scale hierarchical image database," in Proc. IEEE Conf. Comput. Vis. Pattern Recognit. (CVPR), 2009, pp. 248-255.

\bibitem{b16}
T.-Y. Lin, M. Maire, S. Belongie, J. Hays, P. Perona, D. Ramanan, P. Dollár, and C. L. Zitnick, "Microsoft COCO: Common objects in context," in Proc. Eur. Conf. Comput. Vis. (ECCV), 2014, pp. 740-755.

\bibitem{b17}
S. Antol, A. Agrawal, J. Lu, M. Mitchell, D. Batra, C. L. Zitnick, and D. Parikh, "VQA: Visual Question Answering," in Proc. IEEE Int. Conf. Comput. Vis. (ICCV), 2015, pp. 2425-2433.

\bibitem{b18}
 B. M. Lake, T. D. Ullman, J. B. Tenenbaum, and S. J. Gershman, "Building machines that learn and think like people," Behavioral and Brain Sciences, vol. 40, pp. 1-72, 2017.

\bibitem{b19}
W. Schneider and R. M. Shiffrin, "Controlled and automatic human information processing: I. Detection, search, and attention," Psychological Review, vol. 84, no. 1, pp. 1-66, 1977.

\bibitem{b20}
M. Corbetta and G. L. Shulman, "Control of goal-directed and stimulus-driven attention in the brain," Nature Reviews Neuroscience, vol. 3, no. 3, pp. 201-215, 2002.

\bibitem{b21}
C. C. Kemp, A. Edsinger, and E. Torres-Jara, "Challenges for robot manipulation in human environments," IEEE Robotics and Automation Magazine, vol. 14, no. 1, pp. 20-29, 2007.

\bibitem{b22}
D. A. Norman, The Psychology of Everyday Things. New York, NY, USA: Basic Books, 1988.

\end{thebibliography}
\end{document}